# Path Based Mapping Technique for Robots

Amiraj Dhawan[1], Parag Oak, Rahul Mishra, George Puthanpurackal[2]

Department of Computer Engineering,

Fr. C. Rodrigues Institute of Technology, Vashi,

Navi Mumbai, Maharashtra, India

*Abstract*—**The purpose of this paper is to explore a new way of autonomous mapping. Current systems using perception techniques like LAZER or SONAR use probabilistic methods and have a drawback of allowing considerable uncertainty in the mapping process. Our approach is to break down the environment, specifically indoor, into reachable areas and objects, separated by boundaries, and identifying their shape, to render various navigable paths around them. This is a novel method to do away with uncertainties, as far as possible, at the cost of temporal efficiency. Also this system demands only minimum and cheap hardware, as it relies on only Infra-Red sensors to do the job.**

## I. INTRODUCTION

As the research in robotics continues and has become more involved in the past decade, the need for an intelligent navigation system has been realized. Many students and researchers around the world have devoted great amounts of time in developing new ways for an artificial agent to navigate both locally and globally. It has become essential for an agent to be able to navigate unknown environments autonomously. Hence when we talk about navigation we imply three things

• Mapping (the new environment)

• Path Finding

• Actuation

Several methods have been devised but most of them require expensive perception techniques using laser (LIDAR), SONAR or expensive cameras. Such methods are justified when time is crucial or when the environment is highly stochastic. But for simple indoor navigation something simpler can be sufficient.

The maps generated by existing mapping methods are usually discrete and hence can be made more informative and less stochastic either by increasing the resolution of the sensor or by increasing the input sampling rate. Unfortunately, this either increases the cost or increases the processing time and the storage space required.

A desired system would be the one that renders the required map resolution without unreasonable rise in resource consumption. We propose a method that will help an artificial agent map its immediate surroundings in an indoor setting. This method can work with cheap sensors (IR) and will generate a map which is technically discrete but logically continuous. It will allow motion planning with continuous paths.

The proposed system is such that irrespective of the environment complexity or the map resolution, the memory space requirement does not vary drastically and the sensor used will still be the same. The only dimension affected is time and that too varies only with environment complexity while it remains independent of the map resolution. The main steps involved are:

• Generation of Boundaries: The unknown environment is mapped and the boundaries are generated. As a result reachable and unreachable areas are demarcated. Also the detected objects are classified.

• Shape identification: The generated boundaries are analyzed. It is determined whether they are concave or convex in shape. This will supplement the next step.

• Preparing path plan: Finally, continuous paths are generated such that they occupy only the reachable areas of the environment. Path generation depends on the boundary shape.

The layout of the paper is as follows: Section II reviews few existing systems, followed by the proposed system in Section III. The proposed system is broken down into several sections explaining the environment, hardware, boundary generation, shape identification of the boundary, path planning etc. Section IV discusses several merits of the proposed system. Finally we conclude with Section V.

## II. EXISTING SYSTEM

Robotic Mapping is a branch in Robotics, dealing with the application and study of map or floor plan construction by an autonomous robot. There are two types of internal representation:

•Metric: The metric framework is the most common for humans. It considers a 2D space in which it places the objects at their known coordinates. This model is very useful, but is sensitive to noise. Calculation of precise distance is also quite difficult.

•Topological: The topological framework only considers places and their relation. Usually, the metric stored is the distance separating the places. Finally a graph is created in which the nodes correspond to places and the arcs correspond to the paths. Some of the existing systems related to robot mapping are as follows:

### A. Robot Localization and Mapping Using Sonar Data

Without any prior knowledge of the environment the robot generates a global map dynamically and computes the robot location for localization by correlating it with the local map. To create the local map, the robot uses range measurements in different directions from the sonar sensor. A servo holds the sonar sensor, so that 180 degree sweeps are possible [1]. This system uses 2-dimensional grid to provide a map of the robot's environment. Each grid stores the occupancy and certainty value obtained from the robot mapping algorithm which is later used for its localization.

### B. Occupancy Algorithm

The system uses occupancy algorithm to create the map. As the robot explores its environment over time, it uses its range of range sweeping sensor values and current location to determine the occupancy of each grid. It classifies every grid as occupied, empty or unexplored. The occupancy of each cell in the pie is finally updated based on the previous value and the one inferred from the range readings which contributes to the generation of the grip map.

### C. Autonomous topological modeling of an environment

This system uses the method of autonomous topological modeling and localization in a home environment using SONAR. The topological model is extracted from a grid map using cell decomposition and normalized graph cut. Autonomous topological modeling involves the incremental extraction of a sub-region without predefining the number of sub-regions [2]. The following are the important steps involved:

• Cell decomposition can systematically extract empty regions in the grid map and produce a roughly modeled graph structure for an empty environment.

• Normalized graph cut produces an effective clustering result by maximizing the similarity within clusters; this has low computational burden because of the cell decomposition process.

• Finally the topological map is constructed with an unknown number of sub-regions.

### D. Robot Map Creation Algorithm using sensor data

This system describes an algorithm by which a robot can construct a map on the fly, and localize itself to the self-constructed map. In the given system, the robot begins by taking sonar readings, to generate a polar distance map of the robot's immediate neighborhood. These initial soundings are taken to be the robot's initial map [3].

Then the robot starts to move in some direction, stops at a particular location, and takes another sounding. The assumption is taken that there is no major changes in robot's environment, which contributes the best fit sounding map. The best fit returns a most likely location of the robot relative to the origin. The soundings are then shifted with respect to the robot's current location and used to modify the map. Several iterations of this cycle are performed until the robot has finished exploring.

### III. PROPOSED SYSTEM

We plan to implement a system for construction of a navigation map and mapping the position and shape of the objects present in an area using a robot. The system would be limited to finite indoor geographical locations. The robot moves around in the enclosed area and at the end gives a map of the area reachable by the robot and the position of the objects present in the area with their estimated shape and size.

### A. Environment

The environment needs to be a finite indoor geographical location like room or courtyards which are surrounded by some form of boundary (like walls). There can be two types of objects:

• Objects which are not in contact with the boundary of the area i.e. Extrinsic Objects

• Objects which are in contact with the boundary of the area i.e. Intrinsic Objects

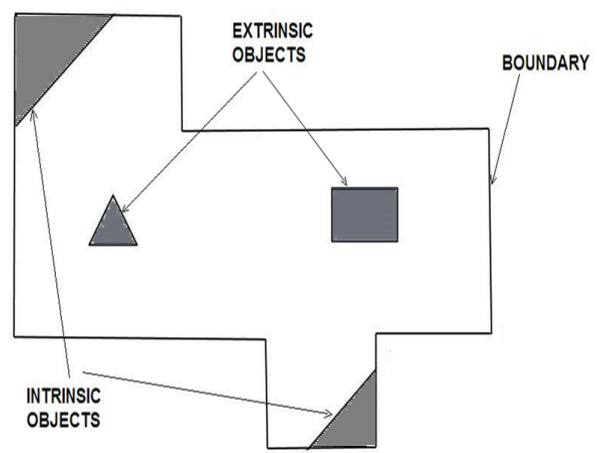

Fig.1.     Boundary generated for Intrinsic and Extrinsic objects.

Intrinsic objects are considered as part of the boundary. Hence the system maps only the reachable area, the boundary of which can be the outline of an Intrinsic Object. For extrinsic objects, the system can report their positions and shapes accurately. Figure 1 shows these two types of objects.

### B. Robot Hardware

The robot will consist of a four wheel or two wheel drive chassis with geared DC motors. The detection of the walls or boundaries can be done by using Digital Infrared Proximity Sensor's on all four sides of the robot. The sensors only detect the presence of the wall at a fixed distance d. More such sensors can be used on the diagonals to increase the mapping quality of the curved boundaries of the room or courtyards. In addition to infrared proximity sensors, the robot would also use wheel encoders to measure the distance traveled in a given time frame.

A Gyroscope sensor would be used to maintain orientation of the robot. The robot will be wireless capable for communication with a host system. The host system would collect data acquired by the robot's sensors and execute the resource heavy algorithm. The host system controls the actions of the robot using a command set. The host system can also be built on the robot itself if the mobile system can provide sufficient computing power for the algorithm to be executed in real time.

### C. Generation of Boundaries

The first task is to map the reachable area by the robot which gives the boundaries of the area and the intrinsic objects. The intrinsic objects and the boundaries are indistinguishable.

The robot starts inside the closed area and randomly moves in any one direction until its proximity sensors give a high signal indicating an obstacle. This obstacle can be:

- An Intrinsic object or the boundary of the area: In which case the robot just moves along the detected obstacle and tries to alter its path whenever the sensor stops detecting the obstacle until the obstacle is again detected by the sensor. For example if the left side sensor detects an obstacle, and on moving along it, the sensor losses the obstacle, the robot would keep moving left, until the sensor again picks up the lost obstacle. The wheel encoders' & proximity sensors' data are sent to the host which checks if the robot has reached the same point from which it first picked up the obstacle signal. At this point the host sends a STOP command to the robot and generation of the boundaries of the area is done. Figure 2 below illustrates this scenario.

- An Extrinsic object: In which case the robot again moves along the detected obstacles and follows the same procedure as in Intrinsic or boundary scenario. But after the boundary generation, the robot finds itself outside the last boundary formed. Since it is a closed finite environment and the robot starts from inside the closed area, this is an impossible case and thus the obstacle found has to be an extrinsic object. Figure 3 shows the mapped extrinsic object.

Now the robot starts again by selecting a random direction and repeats the process but this time ignoring all the previous extrinsic objects mapped.

The boundary generation phase repeats every time when the extrinsic object scenario is detected and it ends successfully as soon as the system detects that the mapped obstacle is an intrinsic object or a boundary scenario.

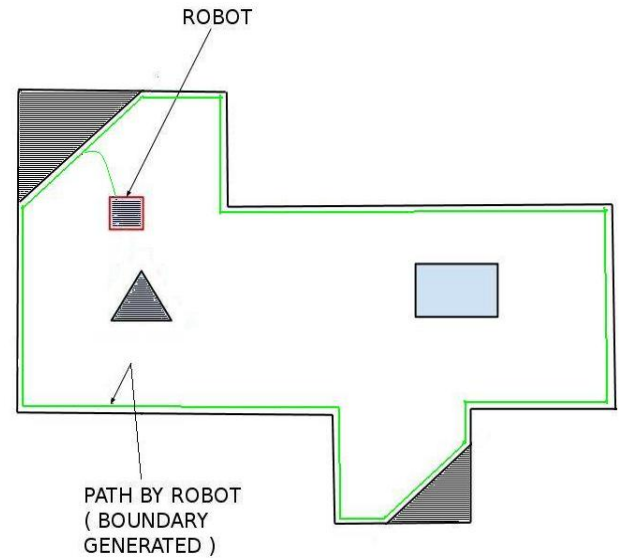

Fig.2.    Generation of Boundary of an Intrinsic Object

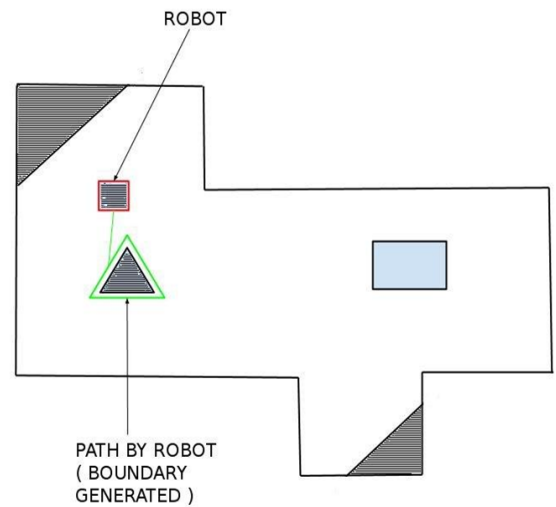

Fig.3.    Generation of Boundary phase for Extrinsic Object Scenario

### D. Shape Identification of the Generated Boundary

After the generation of the boundary, we try to identify the shape of the boundary. This phase is important since different strategies are applied to boundaries depending on its shape.

The types of shapes that are to be identified are Concave Polygon, Convex Polygon, Circular and Complex shapes

. The complex shape is basically a mixture of Convex/Concave Polygon and Circular arcs.

The first step is to detect all the points where the robot took a turn of more than 1 degree in any direction. If no such points are detected then the shape of the generated boundary is Circular. Using these points, the shape can be identified as Concave or Convex Polygon easily using one of various Convex Hull algorithms with slight variations [4]. The preferred algorithm is Jarvis's March also known as Gift Wrapping algorithm due to its running time of O(n) where n is the number of points [5]. According to Jarvis's March algorithm, if at any time the succeeding point vector goes in an anti-clockwise direction then the boundary is classified as a concave polygon and convex polygon otherwise. A complex shape also is passed as either a concave or a convex polygon. For this, the generated boundary is passed through Hough Transform for Circle. If a circle or curve is detected by the algorithm then it is a complex shape. In case a complex shape is detected, the shape is divided into convex/concave polygon and a circular part. These parts are then treated as individual reachable areas by the rest of the algorithm and the objects are mapped independently in these parts.

### E. Preparing The Path Plan

To map the extrinsic objects, we first create a Path Plan using which the robot decides the movement in the reachable area. The path plan is basically a set of straight lines starting from one of the vertices of the reachable area to the edge opposite to the vertex. It also contains the angle at which the robot should start moving from the starting point of the line. We define a parameter which defines how accurately extrinsic objects are detected. This accuracy factor is denoted by the variable $\alpha$. The accuracy factor $\alpha$ can help us deduce the approximate dimensions of the extrinsic objects which might not be mapped by the robot. Thus by customizing the accuracy factor, it would be possible to alter the completeness in the mapping of extrinsic objects. Also, increasing the accuracy would increase the time required for the mapping. We use different approaches for the different types of shapes of the reachable area mapped by the robot as follows:

- *Convex Shape*

The host system first finds all the vertices of the convex shape. Then from the first vertex a straight line is formed with any vertex of the opposite edge. This straight line is added into the path plan for the robot. Figure 4 shows the vertex P of a convex reachable area.

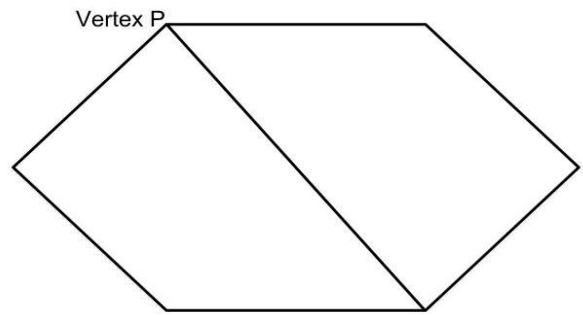

Fig.4.    First path formed from Vertex P

Now we recursively form such straight lines from the same vertex P towards the opposite edges until we reach the other end vertex say R of the opposite edge. The endpoint of each straight line is at a distance $\alpha$ apart from the previous straight line. If the opposite edge length is not a multiple of $\alpha$, then the last straight line is at a distance less then $\alpha$ from the straight line between P and R maintaining the accuracy of $\alpha$.

From the reachable area we can deduce the vertical distance between the vertex P and the opposite edge. Using this vertical distance and $\alpha$, we calculate the angle that each straight line makes with the opposite edge and store them in the path plan.

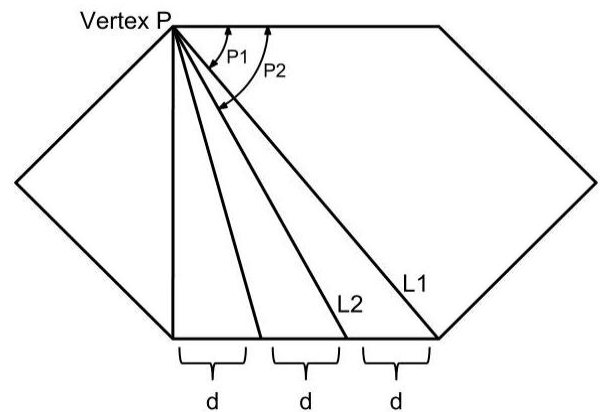

Fig.5.    Straight lines from vertex P to the current opposite edge

We now take another edge opposite to the vertex P, if any, and redo the process considering this new edge. Figure 6 shows all the straight lines from the vertex P.

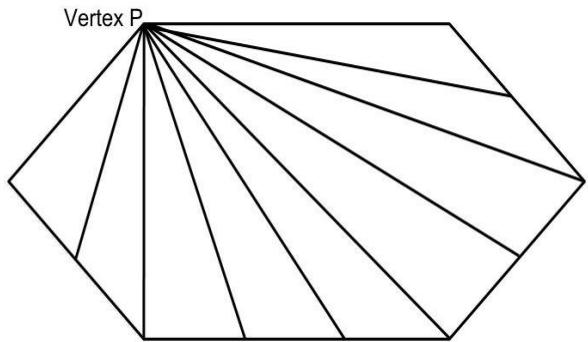

Fig.6.        All the straight lines from vertex P towards the opposite edge

A sample entry in the path plan would consist of the starting vertex, angle made by the line to the boundary in the anti-clockwise direction and a line identifier. For example [P, P1, L1] and [P, P2, L2]. The path plan is complete when the above procedure is repeated for all the vertices. Straight lines in the completed path plan that coincides with any of the reachable area boundary are excluded from the path plan since the path was already used by the robot for mapping the reachable boundary. Figure 7 shows the complete path plan of the example convex shape.

• Concave Shape

For concave shape, we do the same procedure as we did in the convex shape with one addition. Every time a straight line is added into the path plan, it is first checked if the entire line falls with the boundary or not. If any part of the line is outside the reachable area i.e. the concave shape, it is not added into the path plan. Figure 8 shows the accepted and rejected straight lines from the Vertex P.

After every vertex is done, the final Path Plan for the sample Concave Shape is shown in Figure 9.

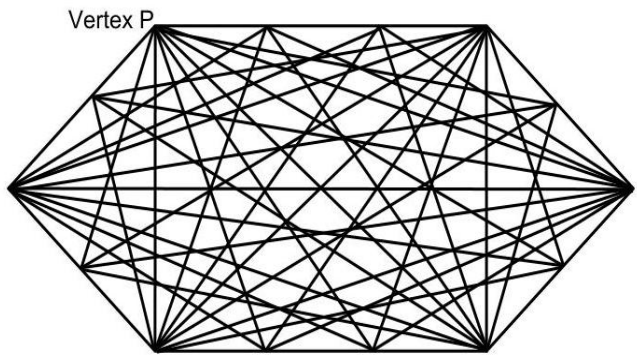

Fig.7.        Path Plan for the sample convex shape

• *Circular Shape*

For Circular shape reachable area, the host system starts from a random point on the circular shape and makes a straight line passing from the center. The next point on the circular shape is selected at an arc length of α from the last point in a clockwise direction. The process ends when the new point has already been processed. If the circular shape is a semicircle, then the straight lines start from the boundary and ends at the center. But in a full circle, the straight lines start the boundary, passes through the center and ends on the other side of the boundary. Figure 10 shows the complete path plan for a circle shape [6].

*F. Mapping an Extrinsic Object*

After preparing the path plan, the host system commands the robot to move according to this plan. At all times, the proximity sensors are active and sends a stop signal to the robot if an obstacle is detected. The vertex, say C, closest to the robot's current position after mapping the boundary and getting the reachable area, is first selected, and all the paths stored in the path plan with starting vertex C are traversed. The path with the lowest angle is selected first and after reaching the opposite edge, the next path, which is closest to the first path, is traversed from the opposite edge to the vertex. During traversing on any path, if any of the proximity sensors send a Stop signal, the robot sends back the information to the host system indicating that it has detected an extrinsic object. The host system, who knows the current position of the robot, the position of the boundary and the direction of the proximity sensor, act differently as follows:

- The robot is sufficiently close to the boundary. In this case the robot is commanded to continue its path without any deviation.

- An extrinsic object, which the host system is already aware of, is detected. In this case, the robot is commanded to ignore the object and alter the path to move along the boundary of the object until it reaches its original path and then continues.

- Detects an extrinsic object, which the host system is unaware of. In this case the host system creates an entry of the extrinsic object. The robot is commanded to alter its path to move along the boundary of the object. It moves in the anti-clockwise direction, along the object if the right or the center proximity sensors detected the object else it moves clockwise. After moving along the object if the robot comes back on its actual path which it was following, the robot stops altering its path according to the object and continues along the original path, reaching the opposite edge of the starting vertex. Then it again takes the exact same path from the path plan but this time starts from the opposite edge and towards the vertex.

Again it picks up the same object and it alters its path along the boundary of this object going along the same side as it did in the previous traversal. The movement data from the wheel encoders and gyroscope data are used by the host system to map the extrinsic object detected, into the reachable area. After reaching the starting vertex, the entire object would be mapped since the robot went around the object for the entire 360 degrees during the 2 traverses of the same path.

When all of the paths in the path plan are traversed at least once, the mapping system ends. The Host system gives an image of the reachable area boundary, and the mapped Extrinsic objects in the environment.

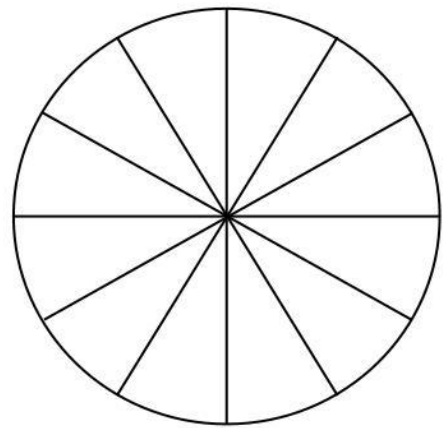

Fig.10.    Sample path plan for a circle

## IV.    COMPARISION WITH EXISTING SYSTEM

One major advantage of our proposed system over existing ones is our system uses IR technology unlike SONAR, LASER technique as used by others. Besides this some points worth noting are:

*1)    The storage space required by the map does not vary too much with change in the accuracy factor unlike current systems using LASER or SONAR. On the other hand, increasing accuracy of any of the existing systems, demands better sampling rates or higher resolution, hence requiring more storage and processing resources. The complexity in processing or using the map generated by proposed system remains independent of the accuracy.*

*2)    Since the proposed system does not use any SONAR sensor, that use audio frequencies, to detect objects, the drawbacks posed by the existing SONAR technology such as with textured walls are not an issue. Also in case of mapping of an auditorium kind of a structure where sound gets absorbed and reflected in a particular way, the accuracy of the SONAR technology worsens, whereas in our proposed system the accuracy would still remain intact.*

*3)    Mapping in areas like courtyards without any roofs is not possible using the SONAR technology since the sound wave loss is far too much. But our proposed system can be used in such conditions and areas since IR technology is unidirectional and the loss is minimal.*

*4)    The object detection by the proposed system is robust. That is the user can be sure that an object of particular dimensions would be recognized by the robot, while in case of the current sound based systems, object detection depends on the extent of its reflective nature and probabilistic models used.*

*5)    Extending the above point, the accuracy factor α described in the proposed system gives a measure to the dimension of the objects that would be detected by the robot. For example, if the robot is mapping in an area where a lot of small objects are expected to be encountered, the accuracy factor can be set to suit the environment i.e. detect the smaller*

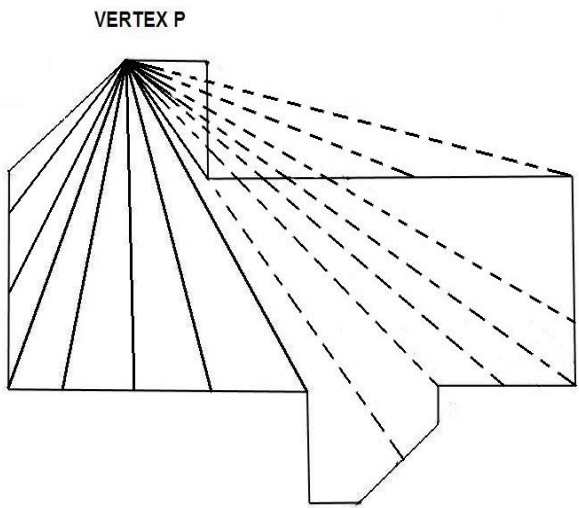

Fig.8.    Accepted and rejected straight lines from Vertex P for a Concave shaped boundary

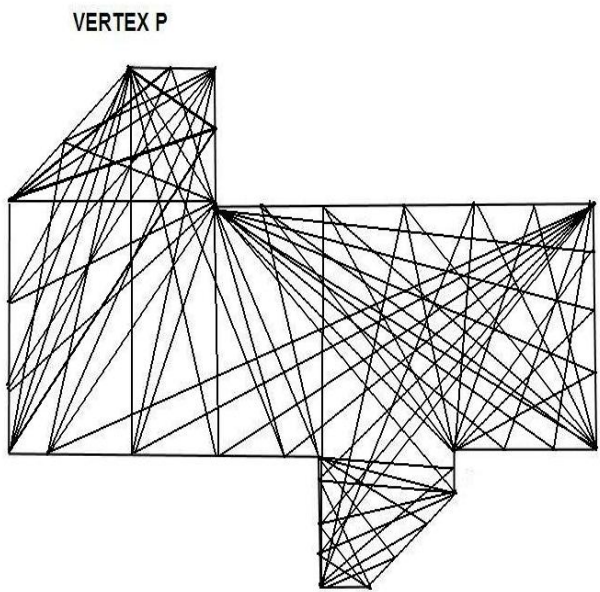

Fig.9.    Shows the final path plan for the sample Concave boundary

objects. Thus changing the accuracy factor allows the robot to behave optimally in different surroundings.

6) *The proposed system as compared to the existing system also gives more focus on detecting objects present at corners, this is due to the fact that one of the endpoint of each path line is a corner, so they have a denser network of path lines thus they provide more precision for mapping objects. Hence smaller objects present near the corners have better chances of detection, as compared to one in the interiors. It has applications in areas like autonomous cleaning robots where corners need to be emphasized.*

## V. CONCLUSION

The gist of the entire technique is as follows:

### A. Generation of the boundaries:

The outermost boundary is mapped and objects within it are also detected. They are classified as follows:

- Extrinsic - these are isolated objects. Only some or none of the objects may be detected at this stage. The remaining ones are mapped after the path generation

- Intrinsic - these are objects attached to the outermost boundary and not differentiated from it.

### B. Shape Identification of the Generated Boundary:

It is determined whether the boundaries are concave, convex, circular or complex. For this the turns made by the robot if greater than 1 degree are recorded as points.

- If there are no such points on a boundary, then it is a circle

- Else Jarvis March Algorithm is used to determine whether it is concave or convex boundary

- Complex boundaries are broken into their simpler components i.e. concave polygon or convex polygon and a circle (detected using Hough Transform)

### C. Preparing the Path Plan:

Depending on the boundary shape the following steps are applied:

- Convex shape - For this the vertices and the edges (at distances determined by accuracy factor α) are connected to form paths.

- Concave shape - The approach is same as that for convex but requires removing the paths that do not lie entirely within the boundary.

- Circular shape - Multiple diameters act as paths. Each adjacent diameter is spaced apart by an angle determined by accuracy factor α.

### D. Mapping extrinsic objects:

Here we are only mapping the objects that were not covered in the first step. The paths that have been generated are traversed. Whenever objects are encountered for the first time they are mapped.

We must keep in mind that the effectiveness of this technique will vary with the type of environment. For example more number of corners will require more paths to be traversed, in turn increasing time required for mapping.

The variable accuracy factor α is the essence of this technique, but it needs to be used carefully. With increase in the accuracy, i.e. with decrease in value of α, the number of paths traversed will increase. This means the system will require more time to complete the mapping process. The time required is directly proportional to the number of paths to be traversed.

We are trying to make major improvements by finding out ways to eliminate paths during planning, mainly redundant paths. As we eliminate more paths, the time required to map will also improve. This will make the technique useful in a wider range of environments.

## ACKNOWLEDGMENT

We would like to express our sincere gratitude to Prof. H.K. Kaura, the Head of Department, Computer Engineering, our guide Ms. M. Kiruthika and Ms. Smita Dange for their support and guidance. Also we thank our friend Rohit Jha for his invaluable views.